%% file: acl_latex.tex
\documentclass[11pt]{article}

\usepackage[final]{acl}

\usepackage{times}
\usepackage{latexsym}
\usepackage{subcaption}

\usepackage[T1]{fontenc}
\usepackage{amsmath}
\usepackage[utf8]{inputenc}

\usepackage{microtype}

\usepackage{inconsolata}

\usepackage{graphicx}
\usepackage{comment}
\usepackage{booktabs}
\usepackage{multirow}
\usepackage{array}
\usepackage{enumitem}
\usepackage{algorithm}
\usepackage{algpseudocode}
\usepackage{xspace}
\usepackage{tabularx}

\newcommand\ours{\texttt{\textbf{BRT}}\xspace}

\title{Block Recursive Transformers for Structured Parameter Sharing}

\author{
  \textbf{Josep Marques\textsuperscript{1}},
  \textbf{Hyunyoung A. Cho\textsuperscript{1}},
  \textbf{Youngeun Kim\textsuperscript{2}},
  \textbf{Youjia Zhang\textsuperscript{1}},
\\
  \textbf{Aecheon Jung\textsuperscript{1}},
  \textbf{Kyuhong Shim\textsuperscript{1}},
  \textbf{Sungeun Hong\textsuperscript{1\textdagger}}
\\
\\
  \textsuperscript{1}Sungkyunkwan University
\\
  \textsuperscript{2}Korea University
}

\begin{document}
\floatstyle{plain}
\restylefloat{algorithm}
\maketitle
\begingroup
\renewcommand{\thefootnote}{\fnsymbol{footnote}}
\footnotetext[2]{
    \fontsize{8.5pt}{10pt}\selectfont
    Corresponding author: Sungeun Hong (\href{mailto:csehong@skku.edu}{csehong@skku.edu})}
\endgroup
\begin{abstract}

Recursive Transformers depend not only on how many parameters are shared, but also on how shared parameters are organized across depth. Existing static schemes adopt two extreme reuse topologies: layer-wise repetition, where each layer is reused independently, and global cyclic reuse, where the entire stack is repeatedly applied as a single shared function. 
We argue that sharing granularity is an important factor in recursive parameter sharing.
This study proposes Block Recursive Transformers (\ours), which partition layers into contiguous blocks and recursively reuse each block before moving to the next. This intermediate topology enables local refinement within blocks while preserving depth-wise transitions, without adding routing modules, auxiliary states, or extra training objectives. Across decoder-only models up to 1.7B parameters, with additional encoder-decoder and encoder-only studies, \ours improves validation negative log-likelihood and downstream accuracy over static recursive sharing baselines. Ablations and representation analyses further show that intermediate block granularity better balances local refinement and depth-wise differentiation. Project website: \url{https://aim-skku.github.io/BRT/}

\end{abstract}

\section{Introduction}

Recursive Transformers reduce the number of unique model parameters by reusing layers while preserving the executed depth of the network \citep{shen2022sliced,baerelaxed,geiping2026scaling}. 
This makes parameter sharing attractive for compact language models, where the goal is to retain the benefits of deep computation while reducing the number of unique trainable layers.

\input{Figures/Figure1}

Existing static sharing strategies follow two extreme reuse topologies. 
Sequence sharing repeats each layer independently before moving to the next, \vspace{-1mm}as in Sequence-style assignment and intra-layer recurrence \citep{takase2023lessons,nguyen2025intra}. 
This preserves a notion of depth, but treats each layer as a separate reuse unit, making it difficult to reuse multi-layer transformations as a coherent stage. 
In contrast, Cycle sharing repeatedly applies the shared stack as a single global function, as in cyclic or recursive reuse schemes \citep{takase2023lessons,baerelaxed,baemixture,yu2025mesh}. 
This improves downstream performance, but forces the same transformation to serve all depth-specific roles, which can cause redundant updates and representational saturation \citep{yu2025mesh}. 
Other compact language model designs use layer sharing for efficient generation \citep{liu2024mobilellm,reid2021subformer}, but do not explicitly study recursive reuse topology as the main design axis. 
As illustrated in Figure~\ref{fig:overview}, Sequence and Cycle sharing form two boundary cases: overly fine-grained reuse and overly global reuse.

This raises the central question of this work: \textit{How should shared parameters be organized across depth?}
Transformer depth is not a homogeneous sequence of interchangeable layers. 
Prior analyses suggest that layers at different depths often play specialized functional roles \citep{liu2025layer,sun2025transformer,shim2022understanding}. 
For example, lower layers tend to preserve more local and lexical information, while upper layers integrate higher-level features and become more closely tied to prediction behavior \citep{zhang2024investigating}. 
This motivates a reuse unit that is larger than a single layer, so that local transformations can be reused as a coherent stage, but smaller than the entire network, so that different depth regions can remain distinct.

In this study, we introduce \emph{Block Recursive Transformers} (\ours), a recursive sharing architecture that partitions depth into contiguous blocks and recursively reuses each block before moving to the next. 
This intermediate topology enables local refinement, repeated application of a short multi-layer transformation within the same depth region, while preserving transitions across depth. 
We add no routing modules \citep{baemixture,yu2025mesh}, auxiliary memory states \citep{yu2025mesh}, layer-wise adaptation modules \citep{baerelaxed}, or extra training objectives. 
By only changing how shared layers are organized and executed, \ours isolates reuse topology as the main variable within a unified static sharing framework.

Our method improves downstream accuracy and validation negative log-likelihood over static recursive sharing baselines, and ablations and representation analyses support the benefit of intermediate block granularity. Rather than adding adaptive mechanisms, BRT isolates how static shared layers should be organized across depth, treating reuse topology as the main design axis for comparing Sequence, Cycle, and block-level reuse within a unified framework.

\input{Figures/Figure2}

Our main contributions are as follows:
\begin{itemize}
    \item We formulate static recursive parameter sharing as a reuse-topology problem and introduce \emph{Block Recursive Transformers} (\ours), which recursively reuse contiguous groups of layers.

    \item We show that Sequence and Cycle sharing are boundary cases of the same block-wise formulation, enabling controlled comparison across reuse granularities without routing modules, auxiliary states, or extra objectives.

    \item We evaluate \ours across decoder-only model sizes, data budgets, and recursion settings, with additional encoder-decoder and encoder-only validation-NLL studies. \ours improves over static recursive sharing baselines with fewer unique trainable parameters.
\end{itemize}

\section{Related Work}

\paragraph{Cross-layer Parameter Sharing.}
Parameter sharing has been widely studied as a way to reduce the number of unique parameters in Transformers.
Early approaches impose strong tying across depth.
Universal Transformers apply the same transformation recurrently across layers \citep{dehghaniuniversal}, while ALBERT shares parameters across Transformer layers to reduce memory usage \citep{lanalbert}.
Although effective for compression, reusing a single layer can limit the representational capacity of different depth regions.

Recent approaches relax this constraint by sharing some components or adding layer-specific capacity.
Subformer combines middle-layer sharing with embedding factorization for generative Transformers \citep{reid2021subformer}, and EdgeFormer augments shared layers with layer-specific adaptation for efficient on-device generation \citep{ge2022edgeformer}.
MobileLLM uses layer sharing for compact sub-billion language models \citep{liu2024mobilellm}, while Basis Sharing decomposes layer weights into shared bases and layer-specific coefficients \citep{wangbasis}.
Overall, these methods mainly study which parameters are shared or how sharing constraints are relaxed.
In contrast, \ours studies how shared parameters are reused after they are tied.
It keeps the layer parameterization unchanged, but reorganizes execution by grouping adjacent layers into recursive blocks.
This shifts the focus from parameter tying itself to the topology of reuse across depth.

\paragraph{Recursive and Adaptive-depth Transformers.}
Recursive Transformers construct depth by repeatedly applying shared layers, making parameter reuse an explicit iterative computation mechanism.
Relaxed Recursive Transformers improve this idea with layer-wise LoRA adaptation \citep{baerelaxed}, while recurrent-depth models study repeated latent computation for test-time scaling \citep{geiping2026scaling}. 
Related looped Transformer studies further show that repeated computation can support iterative refinement and algorithmic behavior \citep{yanglooped,giannou2023looped,fanlooped,gatmiry2024can}. 
Most static recursive sharing schemes, however, rely on either layer-wise repetition or a single global loop, corresponding to Sequence and Cycle sharing strategies \citep{takase2023lessons}. 
These topologies are parameter-efficient, but they either fragment reuse into individual layers or force a single shared stack to serve all depth-specific roles.

Recent adaptive-depth methods introduce mechanisms to decide how much computation each token should receive.
Mixture-of-Recursions learns token-level recursive depths through routing \citep{baemixture}, while MeSH introduces a routing mechanism to support recursive computation \citep{yu2025mesh}.
Unlike these methods, which change the amount or path of computation through routing, memory states, or adaptive schedules, we keep the computation schedule fixed and change only the granularity of reuse.
This makes \ours a controlled setting for studying reuse topology and an orthogonal complement to routing- or memory-based recursive models.

\section{Block Recursive Transformers}
\label{sec:method}

\subsection{Reuse Granularity Analysis}

\ours treats reuse granularity as a controllable design axis.
Instead of reusing individual layers or the full stack, it recursively reuses contiguous groups of layers, reflecting the observation that Transformer layers often form depth-dependent functional groups rather than a uniform stack of interchangeable transformations \citep{geva2021transformer,sun2025transformer,zhang2024investigating}.

To examine how different reuse topologies affect depth-wise representations, we use Centered Kernel Alignment (CKA) \citep{kornblith2019similarity} as a diagnostic tool.
CKA measures the similarity between hidden-state representations, allowing us to compare how representations evolve across layers and recursion steps.
Importantly, CKA is not used to choose block boundaries.
Instead, \ours adopts a fixed contiguous partition, so that the number of blocks directly controls reuse granularity and can be evaluated through ablations.

Figure~\ref{fig:cka_similarity} shows the resulting CKA patterns.
The vanilla Transformer exhibits gradual representation changes across depth, suggesting that adjacent layers often form locally coherent stages.
Sequence sharing shows high similarity across layers, indicating limited depth-wise differentiation when each layer is reused independently.
Cycle sharing shows high similarity across recursion steps, consistent with representational saturation under global reuse \citep{yu2025mesh}.
In contrast, \ours shows higher similarity within each block and lower similarity across blocks.
This pattern supports the view that block-level reuse enables local refinement within each depth region while preserving transitions between different regions.
Additional CKA visualizations are provided in Appendix~\ref{sec:appendix_c}.

\subsection{Formulation}
We start from a standard $L$-layer Transformer \citep{vaswani2017attention}, whose hidden states are updated as:
\begin{equation}
\label{eq:transformer}
h^{\ell}=\mathcal{B}\left(h^{\ell-1};\theta_{\ell}\right),\qquad \ell\in[1, L],
\end{equation}
where $h^0$ is obtained from the token embedding, $\mathcal{B}$ denotes a Transformer block, and $\theta_{\ell}$ denotes the parameters of layer $\ell$.

\paragraph{Block Reuse.}
Given a target executed depth $L$, let $E$ denote the number of non-recursive boundary layers and $R$ the recursion count.
Following prior recursive Transformer settings \citep{baerelaxed,baemixture}, the recurrent core contains $P=(L-E)/R$ unique recurrent layers when exact depth matching is feasible.
The total number of unique layers is then $T=P+E$, and the executed depth is $L=P\cdot R+E$.

We partition the $P$ unique recurrent layers into $S$ contiguous blocks, where $1\leq S\leq P$.
For clarity, we first describe the case where $P$ is divisible by $S$, so each block contains $K=P/S$ layers.
When this divisibility condition is not satisfied, we use the closest feasible contiguous partition: we choose the largest recurrent-core size $P'\leq P$ such that $P'$ is divisible by $S$, set $K=P'/S$, and use $D_{\mathrm{exec}}=P'R+E$ as the executed depth.
This rule keeps block sizes equal, preserves contiguity, and never gives \ours a larger executed depth than the target vanilla model.
We report the resulting executed depth for all configurations in Appendix~\ref{sec:appendix_a}.

For block $s$, we define its computation as:
\begin{equation}
\label{eq:stage}
\mathcal{F}_s
=
\mathcal{B}(\cdot;\theta_{s,K})
\circ \cdots \circ
\mathcal{B}(\cdot;\theta_{s,1}),
\end{equation}
where $\theta_{s,1},\ldots,\theta_{s,K}$ are the parameters of the $K$ layers in block $s$.
The composition follows the original layer order within each block, matching the forward execution order of a standard Transformer.

The block count $S$ controls the granularity of reuse.
As $S$ increases, each reuse unit becomes smaller and the topology approaches layer-wise repetition.
When $S=1$, the recurrent core becomes a single global loop.
Intermediate values $1<S<P$ define \ours.
In this regime, each block is applied $R$ times before representations move to the next block, enabling local refinement within contiguous layer groups while preserving transitions across depth.
\paragraph{Design Rationale.}
We use contiguous blocks to keep recursive refinement local while allowing different depth regions to remain distinct.
Thus, block-level reuse provides an intermediate topology between layer-wise repetition and global cyclic reuse.

By default, we set $E=2$, keeping the first and last layers independent of the recursive core, following prior work \citep{baemixture}.
The overall computation is then:
\begin{equation}
\label{eq:middle-composition}
h_{\text{out}} =
\mathcal{B}\!\left(
\mathcal{F}_{S}^{R}
\Big(
\cdots
\mathcal{F}_{1}^{R}
(\mathcal{B}(h^{0};\theta_{1}))
\Big);
\theta_{L}
\right),
\end{equation}
where $\mathcal{F}_s^{R}$ denotes $R$ repetitions of block $\mathcal{F}_s$, and $\theta_1,\theta_L$ denote the independent boundary layers.

\paragraph{Unifying Static Sharing Topologies.}
With $S$ controlling reuse granularity and $E$ controlling boundary-layer sharing, prior static strategies arise as boundary cases.

\begin{itemize}[leftmargin=1.33em, itemsep=2pt, topsep=2pt]
    \item $S=1$ gives global cyclic reuse, where the recurrent core is reused as one block. With $E=0$, this recovers Cycle sharing; with $E=2$, it corresponds to Middle-Cycle sharing.
    
    \item $S=P$ gives layer-wise repetition, where each recurrent layer is repeated before moving to the next. With $E=0$, this recovers Sequence sharing; with $E=2$, it corresponds to Middle-Sequence sharing.

    \item Intermediate values $1<S<P$ define \ours, interpolating between these extremes through block-level recursive reuse.
\end{itemize}

\vspace{3mm}

\section{Experiments}

We evaluate \ours to answer three questions:
(i) whether block-level reuse improves over static recursive sharing baselines under matched training budgets,
(ii) whether the gains persist across model scale, data scale, and recursion depth, and
(iii) whether the same reuse topology transfers beyond decoder-only architectures.

\subsection{Experimental Setup}

We use LLaMA-based SmolLM models up to 1.7B parameters \citep{touvron2023llama,allal2024smollm}, pretrained from scratch on FineWeb-Edu \citep{penedo2024fineweb}. 
We evaluate ARC, WinoGrande, MMLU, HellaSwag, and PIQA using the Language Model Evaluation Harness \citep{eval-harness}, and report mean performance across three seeds. 
We further evaluate encoder-decoder T5-220M and encoder-only ModernBERT-149M \citep{raffel2020exploring,warner2025smarter}. 
For language-modeling quality, we compute validation negative log-likelihood on 2,000 FineWeb-Edu samples.

Unless otherwise stated, recursive variants are compared under the same recursion count and training-token budget. 
When exact depth matching is infeasible, we use the closest feasible recursive configuration following prior recursive-sharing setups \citep{baerelaxed,baemixture}. 
Reported parameter counts denote unique trainable non-embedding parameters. 
All main decoder-only models use the same optimizer, learning-rate schedule, batch size, and hardware setup; full details are provided in Appendix~\ref{sec:appendix_a}.

\input{Tables/Table1}

\subsection{Main Results}

We first compare \ours with static recursive sharing baselines using SmolLM-360M trained from scratch for 20B tokens.
The baselines include Sequence and Cycle sharing, as well as Middle-Sequence and Middle-Cycle, where the first and last layers are kept independent of the recursive core.
We evaluate two recursion settings, $R=2$ and $R=4$, which reduce the number of unique recurrent layers under matched or closest feasible executed depth.
All models are evaluated on five benchmarks, and we report task-wise accuracy and average performance over three seeds.
Detailed configurations and executed-depth accounting for all recursive variants are provided in Appendix~\ref{sec:appendix_a}.

As shown in Table~\ref{tab:360_20B}, \ours achieves the best average few-shot accuracy among static recursive sharing baselines under comparable or smaller unique-parameter budgets.
With $R=2$, \ours reaches 40.20 average accuracy, making it competitive with the vanilla model while using substantially fewer unique parameters.
It also improves over Middle-Cycle, the strongest recursive baseline in this setting.
With $R=4$, \ours outperforms all recursive baselines and remains close to the vanilla model while using only 78M parameters.
These results suggest that downstream performance depends not only on the number of shared parameters, but also on how reused layers are organized across depth.

\subsection{Scaling with Compute and Data}

\paragraph{Training Compute.}
We evaluate language-modeling quality under different training compute budgets.
Figure~\ref{fig:nll} reports validation negative log-likelihood (NLL) for SmolLM-360M variants with $R=2$ and $R=4$.

\input{Figures/Figure3}
\input{Figures/Figure4}
We compute NLL on 2,000 FineWeb-Edu validation samples and estimate training compute following \citet{baemixture}, based on the scaling-law compute estimate of \citet{kaplan2020scaling}.
Across both recursion settings, \ours{} achieves the lowest validation NLL among static sharing strategies at all evaluated compute budgets.
With $R=2$, \ours improves over both Sequence and Cycle variants, showing that block-level reuse is beneficial even under moderate recursion.
The gains are more pronounced with $R=4$, where stronger compression leads to larger degradation for prior sharing variants.
This trend is consistent with the view that global cyclic reuse can produce redundant updates, while overly fine-grained layer-wise reuse is also suboptimal.
In contrast, \ours provides a more stable quality-compute trade-off by reusing intermediate contiguous blocks.
Together with Table~\ref{tab:360_20B}, these results show that \ours improves both downstream accuracy and validation NLL.
\input{Tables/Table4}

\paragraph{Matched Training Tokens.}
We next evaluate how different sharing strategies scale with training tokens.
Figure~\ref{fig:few_shot} reports mean accuracy over five benchmarks using SmolLM-135M and SmolLM-360M trained with 1B, 5B, 10B, and 20B tokens.

We evaluate both $R=2$ and $R=4$, and report mean performance over three seeds.
For SmolLM-135M, vanilla models remain stronger than recursive variants in some low-data regimes, consistent with prior observations on small recursive models \citep{baerelaxed,baemixture}.
However, \ours continues to benefit from additional training data and improves over the vanilla model at 20B tokens for both recursion settings.
For SmolLM-360M, \ours outperforms static recursive baselines across token budgets and remains competitive with the vanilla model, especially under $R=2$.
Middle-Cycle also improves over Cycle at larger scale, suggesting that keeping the first and last layers independent is beneficial.
Overall, these results suggest that block-level reuse becomes more stable as model size and training budget increase.

\paragraph{Billion-scale Architectures.}
To test whether the gains of \ours extend beyond sub-billion models, we evaluate SmolLM-1.7B recursive architecture variants trained for 20B tokens.
We compare \ours against Middle-Cycle, the strongest recursive baseline in our main experiments, and the vanilla model.
We report task-wise and average performance across three seeds.
In Table~\ref{tab:1.7B_20B_extended}, \ours achieves 41.63 average accuracy, outperforming the Middle-Cycle variant with 41.33 average accuracy.
It also remains competitive with the vanilla model, which achieves 41.68 average accuracy with approximately twice as many trainable parameters.
The largest gain over Middle-Cycle appears on ARC, suggesting that block-level reuse can better preserve downstream performance in this setting.
These results suggest that the benefits of \ours are not limited to sub-billion models and remain effective at billion-scale model sizes.

\subsection{In-depth Analysis}

\input{Tables/Table3}
\input{Figures/Figure5}
We conduct ablation studies to analyze the effect of block granularity in \ours and to verify whether its gains come from the block-level reuse topology itself, rather than solely from adding boundary-independent layers. We pretrain SmolLM-360M from scratch on 10B tokens and evaluate all parameter-sharing variants on five few-shot benchmarks across three seeds. Table~\ref{tab:360_10B} provides the full task-wise results, while Figure~\ref{fig:ablation_blocks1} reports the average accuracy with the standard deviation values across seeds.

\paragraph{Block Granularity.}
We first vary the number of blocks $S$, while keeping the first and last layers independent. As shown in Table~\ref{tab:360_10B} and Figure~\ref{fig:ablation_blocks1}, \ours achieves the best performance when $S=2$, reaching 38.95 average accuracy and outperforming both the vanilla model and recursive sharing baselines. Increasing the number of blocks beyond this point does not further improve performance. Although $S=3$ remains competitive with the vanilla model and outperforms Cycle sharing, $S=4$ degrades downstream accuracy and even falls below the Cycle baseline. These results suggest that \ours benefits from an intermediate block granularity, where each recursive unit is large enough to coordinate local transformations but still avoids the overly global reuse imposed by Cycle sharing ($S=1$).

\paragraph{Fully Recursive Setting.}
We then remove the independent layers and force all layers to participate in the sharing strategy. This setting tests whether the advantage of \ours persists when the model becomes fully recursive. As shown in Table~\ref{tab:360_10B}, \ours with $S=2$ outperforms both Cycle sharing ($S=1$) and Sequence sharing ($S=16$), while remaining competitive with the vanilla model. This result is important because it shows that \ours can recover most of the vanilla model's performance, unlike prior recursive sharing baselines.

In the fully recursive setting, we also observe that finer block partitions are not necessarily better. Increasing the number of blocks beyond two does not improve performance, following the same trend as when including the boundary-independent layers. As the block granularity becomes too fine, \ours approaches a structure similar to Sequence sharing and may inherit its limitations.

Overall, the $S=2$ configuration provides the best balance between global reuse, as in Cycle sharing, and layer-level independent reuse, as in Sequence sharing. The fully recursive setting further shows that the gains of \ours cannot be explained solely by keeping the first and last layers independent. Although independent layers improve performance, block-level recursion remains stronger than prior sharing baselines even when all layers are recursively reused. These findings suggest that even with a smaller number of parameters and sometimes smaller executed depth, \ours provides a strong inductive bias for recursive Transformers.

\subsection{Architecture Generalization}

To examine whether the benefits of \ours extend beyond decoder-only architectures, we evaluate it on encoder-decoder and encoder-only Transformers. We use T5-220M for the encoder-decoder setting and ModernBERT-149M for the encoder-only setting \citep{raffel2020exploring,warner2025smarter}.

For each architecture, we compare \ours against the strongest Cycle and Sequence sharing variants, Middle-Cycle and Middle-Sequence. All models are pretrained from scratch on 10B tokens from FineWeb-Edu \citep{penedo2024fineweb}, and we report the negative log-likelihood on the FineWeb-Edu validation set. Since T5 and ModernBERT differ in architecture and training objective, we analyze them separately rather than comparing their absolute NLL values.
\vspace{-2mm}
\paragraph{Encoder-decoder Architecture.}
We first evaluate T5-220M. Since encoder-decoder models contain two Transformer stacks, we ablate where the recursive structure is applied. We consider three variants: applying recursion to both the encoder and decoder, only to the decoder, and only to the encoder. This allows us to analyze whether \ours is more effective in the bidirectional encoder, the autoregressive decoder, or their combination.

\input{Figures/Figure7}

In Figure~\ref{fig:enc_dec}, \ours achieves the lowest negative log-likelihood in all three settings. 
When both encoder and decoder are recursive, \ours reduces NLL from 5.64 to 5.56. 
The same trend holds when only one stack is recursive, with the best NLL of 5.52 under decoder-only recursion. 
We hypothesize that tying the decoder's cross-attention projection across recursions makes each decoder block act as an iterative source-conditioned refinement step.

\paragraph{Encoder-only Architecture.}
We next evaluate an encoder-only architecture (ModernBERT-149M) to test whether the benefits of \ours also hold under a masked language modeling objective. Unlike T5, ModernBERT consists of a single bidirectional Transformer stack.
In Figure~\ref{fig:enc_dec}, \ours achieves the lowest validation NLL among encoder-only variants, improving over both Middle-Sequence and Middle-Cycle. This result indicates that the advantage of \ours is not limited to autoregressive decoding or encoder-decoder attention. Even in a fully bidirectional encoder, organizing parameter reuse into blocks provides a more effective recursive structure than either repeating individual layers or cycling through the entire stack.

Overall, these results suggest that block-level reuse is not specific to decoder-only models. Across encoder-decoder and encoder-only architectures, \ours consistently improves over prior static sharing baselines, supporting block granularity and reuse topology as important design factors for recursive Transformers.

\section{Conclusion}

We introduced Block Recursive Transformers (\ours), a static recursive sharing architecture that treats reuse topology as a key design factor. The proposed method partitions layers into contiguous blocks and recursively reuses each block before transitioning to the next, providing an intermediate topology between layer-wise repetition and global cyclic reuse. 
Across model scales, data budgets, recursion settings, and architecture types, \ours improves validation negative log-likelihood and downstream accuracy over static recursive sharing baselines. 
Ablations and representation analyses further show that intermediate block granularity better balances local refinement and depth-wise differentiation. 
More broadly, our findings suggest that the Recursive Transformer community should consider not only how much sharing is used, but also how reuse is organized across depth.

\section*{Limitations} 
Our analysis uses a fixed block partition to study block-level recursion in a controlled setting. We do not examine flexible partitioning strategies, such as asymmetric block designs or learned block partitioning. Investigating adaptive block structures across model scales and architectures is a future direction.
In addition, our task coverage is limited to five few-shot evaluation benchmarks. Evaluating \ours on broader tasks, including multilingual settings, coding, and complex reasoning, is another future direction.

\newpage
\section*{Acknowledgements}
This work was supported by the National Research Foundation (NRF) grant funded by the Korean government (MSIT) (RS-2026-25477177). This research was supported by the MSIT (Ministry of Science and ICT), Korea, under the Graduate School of Virtual Convergence support program (IITP-2026-RS-2023-00254129, 33\%) and the ICAN (ICT Challenge and Advanced Network of HRD) support program (IITP-2026-RS-2024-00436934, 33\%), supervised by the IITP (Institute for Information \& Communications Technology Planning \& Evaluation).
\bibliography{latex/custom}

\clearpage
\appendix
\section*{Appendix}

This appendix provides additional implementation details and supporting analyses for the experiments included in the main paper. We include the model configurations used across architectures, describe the training and evaluation settings, and specify how executed depth and parameter counts are computed for each sharing strategy. We also provide additional visualization results to complement the CKA analysis of the main paper, and include the training dynamics of different sharing strategies. We organize these points as follows:

\begin{itemize}
    \item Appendix~\ref{sec:appendix_a} provides the full experimental setup for the experiments across decoder-only, encoder-only and encoder-decoder architectures, details of the different configurations for the parameter-sharing strategies, and explains how depth is computed for \ours.
    \item Appendix~\ref{sec:appendix_c} extends the CKA visualizations from the main body, including a layer-by-layer visualization of the vanilla and Sequence-sharing strategy, and includes training-dynamic visualizations for \ours, Middle-Cycle and Middle-Sequence.
\end{itemize}
\section{Implementation Details}
\label{sec:appendix_a}

Table~\ref{tab:model-configs} summarizes the five base model configurations used in our experiments. These models cover the three Transformer architecture families considered in this work: decoder-only, encoder-decoder, and encoder-only architectures. For each model we report the original parameter count and executed depth before applying the different recursive parameter sharing strategies. We describe their training setup in the following subsections.
\paragraph{Decoder-only experiments.} For our main experiments, we use a LLaMA-based \citep{touvron2023llama} SmolLM architecture \citep{allal2024smollm}. We evaluate the 135M, 360M, and 1.7B variants to compare parameter-sharing strategies across model scales. Unless otherwise specified, all models are pretrained from scratch on a subset of the FineWeb-Edu dataset \citep{penedo2024fineweb} for up to 20B tokens, depending on the experimental setting. For \ours, we use $S=2$ blocks by default. 

For training, we use 8 NVIDIA A6000 GPUs with a total batch size of 1024. We use AdamW \citep{loshchilovdecoupled} as the optimizer with $\beta_1=0.9$ and $\beta_2=0.95$, and set the peak learning rate to $3\times10^{-3}$. Training is performed using bfloat16 precision, and gradients are clipped to a maximum norm of 1.0. We use a trapezoidal learning rate schedule with a 5\% warmup and 10\% linear cooldown. We set the maximum input sequence length to 2048. 
 
We evaluate the models on five benchmarks from the Language Model Evaluation Harness \citep{eval-harness}: ARC, WinoGrande (WG), MMLU, HellaSwag (HS), and PIQA. We follow standard settings from prior work \citep{sakaguchi2021winogrande,bisk2020piqa,clark2018think,zellers2019hellaswag,hendrycksmeasuring}, and evaluate using 5 shots for all benchmarks except ARC, where 25 shots are used. In addition, we also evaluate the models on the validation set of FineWeb-Edu and report the negative log-likelihood on 2,000 validation samples. 
\input{Tables/models_information}
\paragraph{Encoder-decoder and encoder-only experiments.}

For generalization purposes, we also conduct experiments on encoder-decoder and encoder-only architectures, using T5-220M and ModernBERT-149M, respectively \citep{raffel2020exploring,warner2025smarter}. In both settings, we train recursive variants from scratch for 10B tokens on the FineWeb-Edu dataset, and evaluate validation negative log-likelihood on 2,000 FineWeb-Edu samples \citep{penedo2024fineweb}.

For both architectures, we use AdamW with $\beta_1 = 0.9$ and $\beta_2 = 0.95$, a total batch size of 1024, and 8 NVIDIA A6000 GPUs. Training is performed in bfloat16 precision, and gradients are clipped to a maximum norm of 0.5. We set the peak learning rate to $3 \times 10^{-4}$  and use a trapezoidal learning-rate schedule with a 10\% linear decay.

Unlike the decoder-only experiments, the encoder-decoder experiments use a span-corruption objective. For T5, we use a 20\% warmup, relative position embeddings with maximum distance 1024 and 128 buckets, a decoder target length of 448, a noise density of 15\%, and a mean span length of 3.1. We set the maximum input sequence length to 2048.

For the encoder-only experiments, we use a masked language modeling objective with a 15\% masking rate. ModernBERT uses the same optimizer, peak learning rate, batch size, precision, gradient clipping and learning-rate schedule as the T5 experiments, except that the warmup is reduced to 10\%. The maximum input sequence length is set to 1024.

\input{Tables/pseudocode_depth}
\input{Tables/Table5}
\paragraph{Executed Depth and Total Parameters.} In Algorithm~\ref{alg:brt-depth}, we describe how we calculate executed depth for \ours. As indicated in Section~\ref{sec:method}, it is possible that the division operation is not feasible and that the depths do not match perfectly for all sharing strategies in all configurations. For \ours, we take this into account, and always select the closest depth value possible to the original vanilla depth. In particular, the depth of \ours will never be bigger than the vanilla depth, avoiding any kind of advantage over other baselines. For Sequence, Cycle, and their respective middle variants, we follow \citep{baerelaxed,baemixture}, where sharing strategies are allowed to have executed depths larger than those of the vanilla models. In Table~\ref{tab:param_count} we report the non-embedding parameter count, executed depth, and number of unique layers across different sharing strategies for the different experiments.

\section{Extended Visualization Results}
\label{sec:appendix_c}
\input{Figures/Appendix/vanilla}

In Section~\ref{sec:method}, we motivated Block Recursive Transformers by the hierarchical functional roles that emerge across depth in Transformer models.
Here, we extend our CKA analysis with additional heatmap visualizations to further examine these depth-wise patterns.

As shown in Figure~\ref{fig:vanilla}, the vanilla Transformer exhibits a smooth evolution of representations across layers, with high similarity between adjacent layers and gradual transitions through depth.
This suggests that depth naturally forms a small number of functionally coherent groups rather than a sequence of homogeneous transformations, consistent with our observation that partitioning layers into two blocks improves performance.
In contrast, Sequence sharing shows uniformly high similarity across layers, including between early and late layers (Figure~\ref{fig:heatmap_layerwise}).
This indicates limited depth-wise differentiation, aligning with the performance degradation observed for layer-wise sharing.

In addition, we visualize the behavior of global sharing in Figure~\ref{fig:cka_similarity}. In this setting, all layers are grouped into a single shared function that is repeatedly applied. This design leads to highly redundant representations and limited evolution across recursion steps. One possible interpretation is that the model converges early to a fixed point, after which further recursions provide negligible additional computation. In contrast, \ours appears to better preserve the hierarchical structure of Transformers by exhibiting greater similarity among intra-block layers, with a smooth transition between layers, supporting depth specialization. 
\input{Figures/Appendix/layer_wise}
\paragraph{Training Dynamics.}
\input{Figures/Appendix/Figure_Loss}
Figure~\ref{fig:loss} shows the training dynamics of different sharing strategies using SmolLM-360M trained for 10B tokens. All methods follow a similar trajectory in the early stage of training, suggesting that the initial optimization is comparable across sharing strategies. However, after the initial billion tokens, \ours starts to separate from Middle-Cycle and Middle-Sequence, and achieves consistently lower training loss. This gap becomes more visible as training progresses, especially compared to Middle-Sequence. \ours achieves a training loss of 3.41 at 10B tokens, followed by Middle-Cycle and Middle-Sequence at 3.48 and 3.70, respectively. 

We consider these results important because they show that the advantage of block-level reuse appears during training optimization, not only in final downstream performance. In this setting, \ours uses fewer parameters and smaller executed depth than the corresponding recursive baselines, so the lower training loss cannot be explained by additional capacity or computation. Instead, the results suggest that a better reuse organization could help  optimize the training objective. This also supports the main empirical results obtained in the main paper: intermediate reuse granularity is a more effective sharing strategy relative to cycle and sequence sharing.

Overall, these results position intermediate block granularity as a key design factor for recursive Transformers.

\end{document}

%% file: Figures/Figure1.tex
\begin{figure}[t]
  \includegraphics[width=\columnwidth]{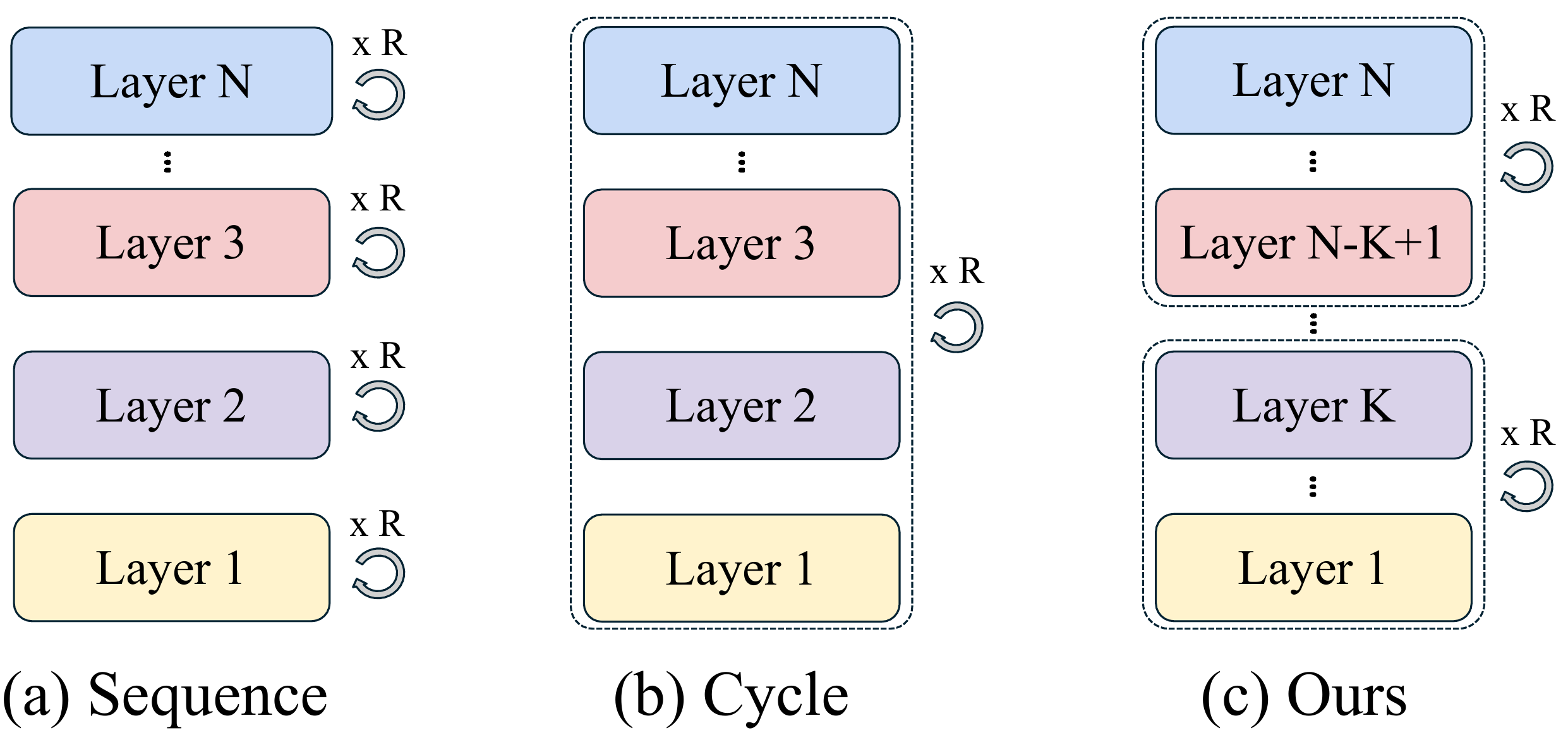}
  \caption{Comparison of sharing strategies. 
(a) Sequence sharing repeats each layer independently.
(b) Cycle sharing reuses the full stack as one global unit.
(c) Our BRT recursively reuses contiguous layer blocks before moving to the next block.
Sequence and Cycle sharing are boundary cases.}
  \label{fig:overview}
\end{figure}

%% file: Figures/Figure2.tex
\begin{figure*}[t]
  \includegraphics[width=1\textwidth]{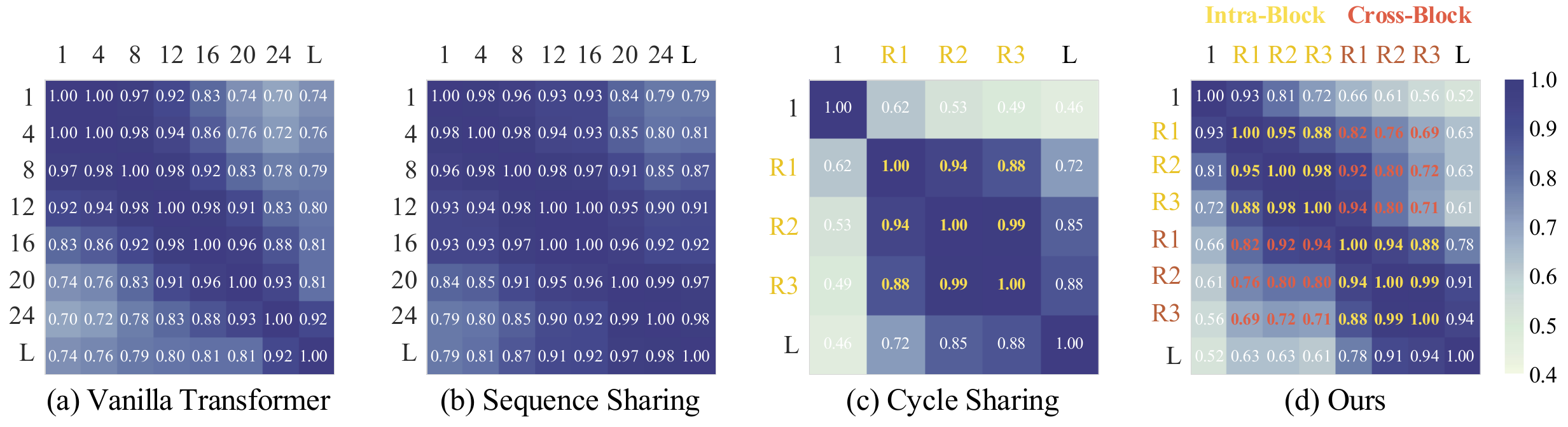}
\caption{
Centered Kernel Alignment (CKA) similarity across depth and recursion steps.
(a) \textit{Vanilla Transformer} shows gradual representation changes across depth.
(b) \textit{Sequence sharing} shows high similarity across layers, suggesting limited depth-wise differentiation.
(c) \textit{Cycle sharing} shows high similarity across recursion steps, consistent with representational saturation under global reuse.
(d) Our BRT shows stronger intra-block similarity and lower cross-block similarity, suggesting local refinement while preserving transitions across depth regions.
}
  \label{fig:cka_similarity}
\end{figure*}

%% file: Tables/Table1.tex
\renewcommand{\arraystretch}{0.85}
\begin{table*}
    \centering
    \setlength{\tabcolsep}{3pt}
    \resizebox{\textwidth}{!}{%
    \begin{tabular}{l| cc rr cccccc}
    \toprule
    \multirow{2}{*}{Model} &
    \multicolumn{2}{c}{Recursion} &
    \multicolumn{2}{c}{Pretrain} &
    \multicolumn{6}{c}{Few-shot Accuracy $\uparrow$} \\
    \cmidrule(lr){2-3} \cmidrule(lr){4-5} \cmidrule(lr){6-11}
    & $N_r$ & Share & Param & $N_{tok}$ & ARC & WG & MMLU & HS & PIQA & Avg \\
    \midrule
    Vanilla & - & - & 315M & 20B & $34.87 \pm 0.31$ & $50.92 \pm 0.41$ & $25.12 \pm 0.17$ & $28.32 \pm 0.16$ & $59.45 \pm 0.32$ & $39.74 \pm 0.14$ \\
    \midrule
    \multirow{10}{*}{Recursive}
    &   & Seq   & 157M & 20B & $32.90 \pm 0.40$ & $50.03 \pm 0.42$ & $24.88 \pm 0.34$ & $27.51 \pm 0.22$ & $59.61 \pm 0.31$ & $38.98 \pm 0.30$ \\
    &   & Cyc   & 157M & 20B & $34.12 \pm 0.50$ & $50.37 \pm 1.13$ & $24.39 \pm 0.26$ & $28.37 \pm 0.10$ & $\textbf{59.90} \pm 0.43$ & $39.43 \pm 0.20$ \\
    & 2 & M-Seq & 167M & 20B & $32.96 \pm 0.11$ & $50.76 \pm 0.34$ & $25.21 \pm 0.41$ & $28.02 \pm 0.34$ & $59.17 \pm 0.50$ & $39.22 \pm 0.23$ \\
    &   & M-Cyc & 167M & 20B & $34.73 \pm 0.21$ & $51.12 \pm 0.63$ & $25.12 \pm 0.19$ & $28.59 \pm 0.12$ & $59.23 \pm 0.10$ & $39.76 \pm 0.12$ \\
    &   & Ours  & 157M & 20B & $\textbf{35.48} \pm 0.35$ & $51.32 \pm 0.64$ & $25.46 \pm 0.22$ & $\textbf{28.98} \pm 0.07$ & $59.74 \pm 0.50$ & $\textbf{40.20} \pm 0.23$ \\
    \cmidrule(l){2-11}
    &   & Seq   & 78M & 20B & $32.03 \pm 0.33$ & $50.83 \pm 1.08$ & $24.38 \pm 0.21$ & $27.34 \pm 0.22$ & $57.88 \pm 0.28$ & $38.49 \pm 0.26$ \\
    &   & Cyc   & 78M & 20B & $33.48 \pm 0.23$ & $50.22 \pm 1.04$ & $\textbf{25.51} \pm 0.18$ & $28.46 \pm 0.10$ & $58.60 \pm 0.35$ & $39.25 \pm 0.13$ \\
    & 4 & M-Seq & 98M & 20B & $32.56 \pm 0.54$ & $51.79 \pm 0.52$ & $24.90 \pm 0.32$ & $27.62 \pm 0.46$ & $58.61 \pm 0.53$ & $39.09 \pm 0.38$ \\
    &   & M-Cyc & 98M & 20B & $33.67 \pm 0.23$ & $51.89 \pm 0.59$ & $25.18 \pm 0.23$ & $28.47 \pm 0.23$ & $59.55 \pm 0.24$ & $39.75 \pm 0.13$ \\
    &   & Ours  & 78M & 20B & $34.43 \pm 0.09$ & $\textbf{51.91} \pm 0.41$ & $25.32 \pm 0.16$ & $28.73 \pm 0.35$ & $59.43 \pm 0.21$ & $39.96 \pm 0.10$ \\
    \bottomrule
    \end{tabular}%
    }
    \caption{
Main results on SmolLM-360M under a fixed 20B-token training budget.
We compare vanilla and recursive Transformers with different static sharing strategies.
$N_r$ denotes the number of recursion steps, and $N_{\text{tok}}$ denotes the number of training tokens.
The best average performance is highlighted in bold.
Results are reported as mean and standard deviation across 3 seeds.
}
    \label{tab:360_20B}
\end{table*}

%% file: Figures/Figure3.tex
\begin{figure}[t]
  \includegraphics[width=\columnwidth]{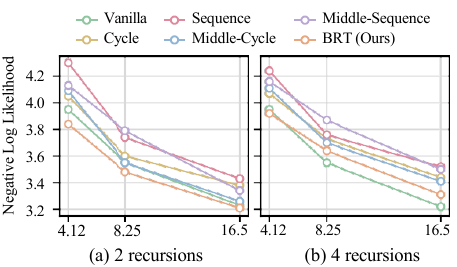}
  \caption{
Validation NLL of SmolLM-360M variants under different training compute budgets. 
The x-axis shows estimated training FLOPs ($\times 10^{18}$). 
NLL is computed on 2,000 FineWeb-Edu validation samples. 
All models are pretrained from scratch; sharing configurations are listed in Table~\ref{tab:param_count}.
}
  \label{fig:nll}
  \vspace{-3mm}
\end{figure}

%% file: Figures/Figure4.tex
\begin{figure*}[t]
\hspace{1.mm}
  \includegraphics[width=0.98\textwidth]{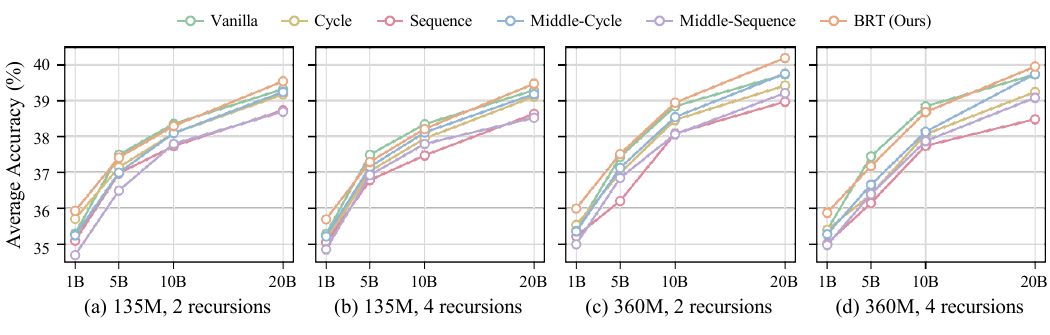}
\caption{
Comparison of parameter-sharing strategies under matched training-token budgets.
We pretrain SmolLM-135M and SmolLM-360M variants with 1B, 5B, 10B, and 20B tokens under two and four recursion settings.
We report mean accuracy over five few-shot benchmarks across three seeds.
}
  \label{fig:few_shot}
\end{figure*}

%% file: Tables/Table4.tex
\begin{table*}
    \centering
    \setlength{\tabcolsep}{3pt}
    \resizebox{\textwidth}{!}{%
    \begin{tabular}{l|cc cc cccccc}
    \toprule
    \multirow{2}{*}{Model} &
    \multicolumn{2}{c}{Recursion} &
    \multicolumn{2}{c}{Pretrain} &
    \multicolumn{6}{c}{Few-shot Accuracy $\uparrow$} \\
    \cmidrule(lr){2-3} \cmidrule(lr){4-5} \cmidrule(lr){6-11}
    & $N_r$ & Share & Param & $N_{tok}$ & ARC & WG & MMLU & HS & PIQA & Avg \\
    \midrule
    Vanilla & - & - & 1.61B & 20B & $38.12 \pm 0.22$ & $\textbf{51.55} \pm 0.72$ & $25.97 \pm 0.26$ & $\textbf{31.76} \pm 0.17$ & $61.03 \pm 0.16$ & $\textbf{41.68} \pm 0.25$ \\
    \midrule
    \multirow{2}{*}{Recursive}
    & 2 & M-Cyc & 870M & 20B & $37.65 \pm 0.26$ & $50.93 \pm 0.89$ & $25.64 \pm 0.21$ & $31.31 \pm 0.32$ & $\textbf{61.13} \pm 0.12$ & $41.33 \pm 0.20$ \\
    & 2 & Ours  & 805M & 20B & $\textbf{38.25} \pm 0.29$ & $51.30 \pm 0.61$ & $\textbf{26.10} \pm 0.18$ & $31.69 \pm 0.22$ & $60.82 \pm 0.27$ & $\underline{41.63} \pm 0.22$ \\
    \bottomrule
    \end{tabular}%
    }
   \caption{Comparison of parameter-sharing strategies in vanilla and recursive Transformers using SmolLM-1.7B. All models are trained for 20B tokens. We report task-wise and mean accuracy across three seeds. The best overall performance is highlighted in bold, while the best recursive strategy is underlined.}

    \label{tab:1.7B_20B_extended}
\end{table*}

%% file: Tables/Table3.tex
\renewcommand{\arraystretch}{0.85}
\begin{table*}
    \centering
    \resizebox{\textwidth}{!}{
    \begin{tabular}{c c l rr cccccc}
    \toprule
    \multicolumn{3}{c}{Sharing} &
    \multicolumn{2}{c}{Pretrain} &
    \multicolumn{6}{c}{Few-shot Accuracy $\uparrow$} \\
    \cmidrule(lr){1-3} \cmidrule(lr){4-5} \cmidrule(lr){6-11}
    $S$ & Untied & Share & Param & $N_{\text{tok}}$ & ARC & WG & MMLU & HS & PIQA & Avg \\
    \midrule

    1  & No  & \ours{} & 157M & 10B & $31.35 \pm 0.39$ & $50.75 \pm 0.59$ & $24.93 \pm 0.26$ & $27.33 \pm 0.13$ & $58.01 \pm 0.16$ & $38.47 \pm 0.24$ \\
    2  & No  & \ours{} & 157M & 10B & $\textbf{32.10} \pm 0.35$ & $50.31 \pm 0.26$ & $24.74 \pm 0.22$ & $27.70 \pm 0.11$ & $\textbf{59.05} \pm 0.75$ & $38.78 \pm 0.24$ \\
    3  & No  & \ours{} & 147M & 10B & $32.08 \pm 0.36$ & $50.31 \pm 0.36$ & $24.53 \pm 0.31$ & $27.54 \pm 0.13$ & $57.92 \pm 0.31$ & $38.47 \pm 0.04$ \\
    4  & No  & \ours{} & 157M & 10B & $32.07 \pm 0.17$ & $49.40 \pm 0.53$ & $24.35 \pm 0.29$ & $27.49 \pm 0.11$ & $58.48 \pm 0.32$ & $38.36 \pm 0.07$ \\
    16 & No  & \ours{} & 157M & 10B & $30.42 \pm 0.50$ & $51.22 \pm 1.17$ & $24.71 \pm 0.19$ & $26.79 \pm 0.20$ & $57.20 \pm 0.20$ & $38.07 \pm 0.32$ \\

    \midrule

    1  & Yes & \ours{} & 167M & 10B & $31.16 \pm 0.20$ & $\textbf{51.85} \pm 1.09$ & $24.70 \pm 0.19$ & $27.39 \pm 0.16$ & $57.65 \pm 0.44$ & $38.55 \pm 0.28$ \\
    2  & Yes & \ours{} & 157M & 10B & $31.95 \pm 0.45$ & $51.57 \pm 0.16$ & $24.84 \pm 0.21$ & $27.47 \pm 0.08$ & $58.94 \pm 0.53$ & $\textbf{38.95} \pm 0.07$ \\
    3  & Yes & \ours{} & 167M & 10B & $31.92 \pm 0.38$ & $50.76 \pm 0.37$ & $\textbf{24.96} \pm 0.36$ & $\textbf{27.78} \pm 0.08$ & $58.44 \pm 0.13$ & $38.77 \pm 0.08$ \\
    4  & Yes & \ours{} & 138M & 10B & $31.47 \pm 0.32$ & $50.60 \pm 0.71$ & $24.68 \pm 0.27$ & $27.23 \pm 0.13$ & $58.13 \pm 0.44$ & $38.42 \pm 0.17$ \\
    15 & Yes & \ours{} & 167M & 10B & $30.31 \pm 0.58$ & $51.28 \pm 1.16$ & $24.23 \pm 0.17$ & $27.37 \pm 0.04$ & $57.18 \pm 0.68$ & $38.07 \pm 0.29$ \\

    \bottomrule
    \end{tabular}
    }
    \caption{
    Block-count ablation of \ours on SmolLM-360M.
    $S$ denotes the number of blocks.
    The \textit{Untied} column indicates whether the first and last layers are excluded from recursive sharing.
    The largest $S$ differs by setting because the recursive core has a different number of layers when boundary layers are untied.
    $S=1$ corresponds to cycle-style sharing, while the largest $S$ values correspond to sequence-style sharing.
    All models are trained for 10B tokens and evaluated on five benchmarks.
    }
    \vspace{-2mm}
    \label{tab:360_10B}
\end{table*}

%% file: Figures/Figure5.tex
\begin{figure}[t]
   \hspace{1mm}
\includegraphics[width=0.92\columnwidth]{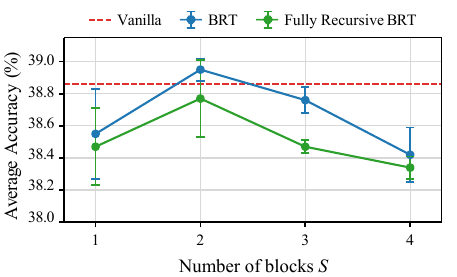}
\caption{
Block-granularity analysis of BRT on SmolLM-360M architecture.
Fully Recursive BRT denotes the setting without independent boundary layers ($E=0$).
All models are trained for 10B tokens.
Points show mean accuracy over five benchmarks and three seeds; error bars show the standard deviation.
}
  \label{fig:ablation_blocks1}
\end{figure}

%% file: Figures/Figure7.tex
\begin{figure}[t]
  \includegraphics[width=\columnwidth]{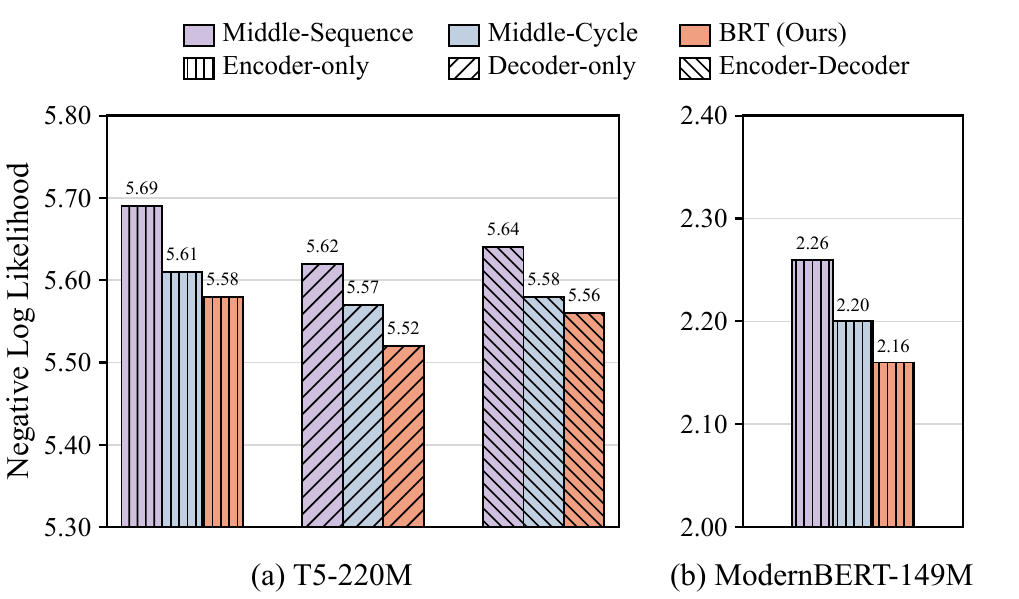}
\caption{
Comparison on encoder-decoder and encoder-only architectures under matched training tokens.
(a) T5-220M with recursion applied to both stacks, decoder only, or encoder only.
(b) Encoder-only ModernBERT-149M.
All models are trained for 10B tokens, and NLL is computed on 2,000 FineWeb-Edu validation samples.
}
  \label{fig:enc_dec}
  \vspace{-1mm}
\end{figure}

%% file: Tables/models_information.tex
\begin{table}[t]
\centering
\small
\setlength{\tabcolsep}{3pt}
\renewcommand{\arraystretch}{1.12}
\begin{tabularx}{\columnwidth}{@{}c|Xcc@{}}
\toprule
& \multicolumn{3}{c}{\textbf{Description}} \\
\cmidrule(lr){2-4}
\textbf{Models} & \textbf{Architecture} & \textbf{Total Params} & \textbf{Layers} \\
\midrule
\multirow[c]{3}{*}{SmolLM}
& Decoder-only & 135M & 30 \\
& Decoder-only & 360M & 32 \\
& Decoder-only & 1.7B & 24 \\
\midrule
T5 & Encoder-decoder & 220M & 12+12 \\
\midrule
ModernBERT & Encoder-only & 149M & 22 \\
\bottomrule
\end{tabularx}
\caption{Description of the models used in our experiments. Total parameters include embedding and non-embedding parameters. The reported values correspond to the original model versions, considering both the embedding and non-embedding parameters.}
\label{tab:model-configs}
\end{table}

%% file: Tables/pseudocode_depth.tex
\floatstyle{ruled}
\restylefloat{algorithm}
\begin{algorithm}[t]
\small
\caption{\textit{BRT Depth Calculation}
}
\label{alg:brt-depth}
\vspace{1mm}
\footnotesize {\textit{Given a target vanilla depth $L$ and a recursive configuration $(E,R,S)$, the procedure returns the largest feasible executed depth $D_{exec} \le L$, together with the number of unique recurrent layers $P$ and layers per block $K$.}}
\vspace{1mm}
\begin{algorithmic}[1]
\State \textbf{Input:} vanilla depth $L$, independent layers $E \in \{0,2\}$, recursions $R \in \{2,4\}$, blocks $S$
\State \textbf{Output:} BRT executed depth $D_{exec}$, unique recurrent layers $P$, layers per block $K$
    \If{$(L-E) \bmod R = 0$}
        \State $P \gets (L-E) / R$
        \Comment{is divisible}
    \Else
        \State $P \gets \lfloor (L-E) / R \rfloor$
        \Comment{select 1 less}
    \EndIf
    \If{$P \bmod S = 0$}
        \State $K \gets P / S$
        \Comment{is divisible}
    \Else
        \State $K \gets \lfloor P / S \rfloor$
        \Comment{select 1 less}
        \State $P \gets S \cdot K$
        \Comment{Adjust $P$}
    \EndIf
    \State $D_{exec} \gets E + R \cdot S \cdot K$
    \Comment{Final executed depth ($\le L$)}
    \State \Return $D_{exec},\; P,\; K$
\end{algorithmic}
\end{algorithm}

%% file: Tables/Table5.tex
\renewcommand{\arraystretch}{0.96}
\begin{table}[t]
    \centering
    \setlength{\tabcolsep}{3pt}
    \resizebox{\columnwidth}{!}{%
    \begin{tabular}{l|l|c|c|c|c|c|c}
    \toprule
    Base Model & Setting & Share & $N_{\text{r}}$ & $S$ & Param & $D_{\text{exec}}$ & $L_{\text{uniq}}$ \\
    \midrule

    \multirow{11}{*}{SmolLM-135M}
        & Vanilla & -- & -- & -- & 106M & 30 & 30 \\
    \noalign{\vskip 1.5pt}
    \cline{2-8}
    \noalign{\vskip 1.5pt}
        & \multirow{10}{*}{Recursive}
        & Seq   & \multirow{5}{*}{2} & 15 & 53M & 30 & 15 \\
        &        & Cyc   &                    & 1  & 53M & 30 & 15 \\
        &        & M-Seq &                    & 14 & 57M & 30 & 16 \\
        &        & M-Cyc &                    & 1  & 57M & 30 & 16 \\
        &        & \ours &                    & 2  & 57M & 30 & 16 \\
    \noalign{\vskip 1.5pt}
    \cline{3-8}
    \noalign{\vskip 1.5pt}
        &        & Seq   & \multirow{5}{*}{4} & 8  & 28M & 32 & 8 \\
        &        & Cyc   &                    & 1  & 28M & 32 & 8 \\
        &        & M-Seq &                    & 7  & 32M & 30 & 9 \\
        &        & M-Cyc &                    & 1  & 32M & 30 & 9 \\
        &        & \ours &                    & 2  & 28M & 26 & 8 \\

    \noalign{\vskip 2pt}
    \midrule\midrule
    \noalign{\vskip 2pt}

    \multirow{16}{*}{SmolLM-360M}
        & Vanilla & -- & -- & -- & 315M & 32 & 32 \\
    \noalign{\vskip 1.5pt}
    \cline{2-8}
    \noalign{\vskip 1.5pt}
        & \multirow{15}{*}{Recursive}
        & Seq   & \multirow{10}{*}{2} & 16 & 157M & 32 & 16 \\
        &        & Cyc   &                     & 1  & 157M & 32 & 16 \\
        &        & M-Seq &                     & 15 & 167M & 32 & 17 \\
        &        & M-Cyc &                     & 1  & 167M & 32 & 17 \\
        &        & \multirow{6}{*}{\ours}
                          &                     & 2  & 157M & 32 & 16 \\
        &        &       &                     & 3  & 147M & 30 & 15 \\
        &        &       &                     & 4  & 157M & 32 & 16 \\
        &        &       &                     & 2  & 157M & 30 & 16 \\
        &        &       &                     & 3  & 167M & 32 & 17 \\
        &        &       &                     & 4  & 138M & 26 & 14 \\
    \noalign{\vskip 1.5pt}
    \cline{3-8}
    \noalign{\vskip 1.5pt}
        &        & Seq   & \multirow{5}{*}{4}  & 8  & 78M & 32 & 8 \\
        &        & Cyc   &                     & 1  & 78M & 32 & 8 \\
        &        & M-Seq &                     & 8  & 98M & 34 & 10 \\
        &        & M-Cyc &                     & 1  & 98M & 34 & 10 \\
        &        & \ours &                     & 2  & 78M & 26 & 8 \\

    \noalign{\vskip 2pt}
    \midrule\midrule
    \noalign{\vskip 2pt}

    \multirow{3}{*}{SmolLM-1.7B}
        & Vanilla & -- & -- & -- & 1.61B & 24 & 24 \\
    \noalign{\vskip 1.5pt}
    \cline{2-8}
    \noalign{\vskip 1.5pt}
        & \multirow{2}{*}{Recursive}
        & M-Cyc & \multirow{2}{*}{2} & 1 & 870M & 24 & 13 \\
        &        & \ours &                    & 2 & 805M & 22 & 12 \\

    \bottomrule
    \end{tabular}%
    }

   \caption{Comparison of parameter count, executed depth, and unique layers for different sharing strategies on SmolLM. Base Model denotes the original SmolLM scale, Setting indicates vanilla or recursive models, Param denotes the non-embedding parameters, $N_{\text{r}}$ denotes the recursion number, $S$ denotes the number of blocks, $D_{\text{exec}}$ denotes executed depth, and $L_{\text{uniq}}$ denotes unique layers.}
    \label{tab:param_count}
\end{table}

%% file: Figures/Appendix/vanilla.tex
\begin{figure}[t]
  \centering
  \includegraphics[width=\columnwidth]{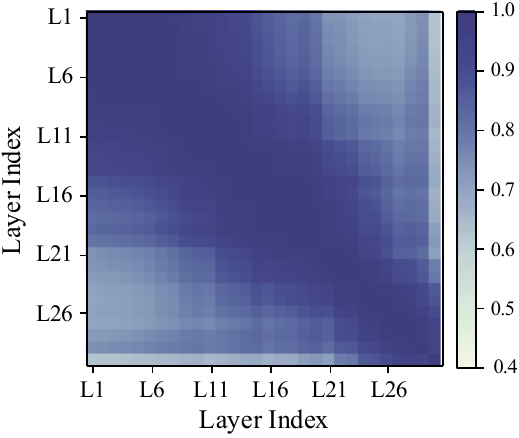}
  \caption{Visualization of the representation similarity across layers in vanilla Transformers.}
  \label{fig:vanilla}
\end{figure}

%% file: Figures/Appendix/layer_wise.tex
\begin{figure}[t]
  \centering
  \includegraphics[width=\columnwidth]{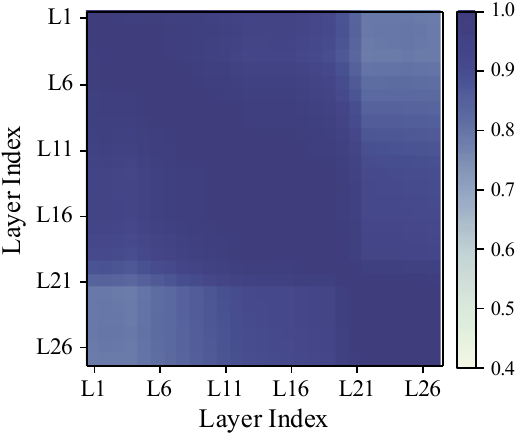}
  \caption{Visualization of the representation similarity across layers in sequence sharing.}
  \label{fig:heatmap_layerwise}
\end{figure}

%% file: Figures/Appendix/Figure_Loss.tex
\begin{figure}[t]
  \centering
  \includegraphics[width=\columnwidth]{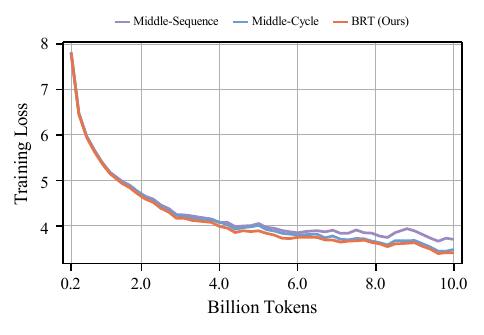}
  \caption{Visualization of the training loss using SmolLM-360M for different sharing strategies. All models are trained for 10B tokens. We set the number of recursions to 2, and the exact executed depth and number of parameters can be found in Table~\ref{tab:param_count}.}
  \label{fig:loss}
\end{figure}